\documentclass{article} 
\usepackage{iclr2024_conference,times}

\usepackage{amsmath,amsfonts,bm}

\def\eqref#1{equation~\ref{#1}}

\def\1{\bm{1}}

\DeclareMathAlphabet{\mathsfit}{\encodingdefault}{\sfdefault}{m}{sl}
\SetMathAlphabet{\mathsfit}{bold}{\encodingdefault}{\sfdefault}{bx}{n}

\usepackage{hyperref}
\usepackage{url}
\usepackage{cleveref}
\usepackage{graphicx}
\usepackage{tabulary,booktabs}
\usepackage{multirow}
\usepackage{subcaption}
\title{Time-Varying Constraint-Aware Reinforcement Learning for Energy Storage Control}

\author{Jaeik Jeong, Tai-Yeon Ku \& Wan-Ki Park \\
Energy ICT Research Section, Electronics and Telecommunications Research Institute (ETRI)\\
Daejeon 34129, Republic of Korea\\
\texttt{\{jaeik1210,kutai,wkpark\}@etri.re.kr} \\
}

\iclrfinalcopy 
\begin{document}

\maketitle

\begin{abstract}
Energy storage devices, such as batteries, thermal energy storages, and hydrogen systems, can help mitigate climate change by ensuring a more stable and sustainable power supply. To maximize the effectiveness of such energy storage, determining the appropriate charging and discharging amounts for each time period is crucial. Reinforcement learning is preferred over traditional optimization for the control of energy storage due to its ability to adapt to dynamic and complex environments. However, the continuous nature of charging and discharging levels in energy storage poses limitations for discrete reinforcement learning, and time-varying feasible charge-discharge range based on  state of charge (SoC) variability also limits the conventional continuous reinforcement learning. In this paper, we propose a continuous reinforcement learning approach that takes into account the time-varying feasible charge-discharge range. An additional objective function was introduced for learning the feasible action range for each time period, supplementing the objectives of training the actor for policy learning and the critic for value learning. This actively promotes the utilization of energy storage by preventing them from getting stuck in suboptimal states, such as continuous full charging or discharging. This is achieved through the enforcement of the charging and discharging levels into the feasible action range. The experimental results demonstrated that the proposed method further maximized the effectiveness of energy storage by actively enhancing its utilization.
\end{abstract}

\section{Introduction}

Energy storage devices, such as batteries, thermal energy storages, and hydrogen systems, play a pivotal role in mitigating the impact of climate change \citep{aneke2016energy,jacob2023future}. These storage technologies are instrumental in capturing and efficiently storing excess energy generated from renewable sources during peak production periods, such as sunny or windy days. By doing so, they enable the strategic release of stored energy during periods of high demand or when renewable energy production is low, thereby optimizing energy distribution and reducing reliance on traditional fossil fuel-based power generation. It can be utilized in energy arbitrage by attempting to charge when surplus energy is generated and energy prices are low or even negative, and conversely, discharging during periods of energy scarcity when prices are high \citep{bradbury2014economic}. The integration of energy storage with arbitrage strategies contributes to grid stability, enhances overall energy reliability, and fosters a more sustainable energy ecosystem. Energy storage devices are utilized in various capacities, ranging from small-scale applications for households to large-scale units for the overall grid operation, contributing to mitigating climate change \citep{ku2022energy,inage2019role}.

To enhance the utility of energy storage devices, determining optimal charge and discharge levels for each time period is crucial. In recent times, reinforcement learning techniques have gained prominence over traditional optimization methods for this purpose \citep{cao2020deep,jeong2023deep}. Unlike conventional optimization approaches, reinforcement learning allows for dynamic adaptation and decision-making in response to changing conditions, enabling energy storage systems to continuously learn and improve their performance over time. This shift towards reinforcement learning reflects a recognition of its ability to navigate complex and dynamic environments. It offers a more adaptive and effective solution to optimize charging and discharging strategies for energy storage devices across diverse temporal patterns. This transition in methodology underscores the importance of leveraging advanced learning algorithms to maximize the operational efficiency of energy storage systems in real-world, time-varying scenarios.

The continuous nature of energy storage device charge and discharge levels poses a challenge when employing discrete reinforcement learning techniques such as Q-learning. These algorithms operate with a predefined set of discrete actions, e.g., fully or partially charging/discharging \citep{cao2020deep,rezaeimozafar2024hybrid}. It limits their suitability for tasks involving continuous variables, and thereby constraining the system's ability to explore and optimize across a continuous range of charge and discharge values. Consequently, the utilization of these methods may lead to suboptimal solutions, as the algorithms cannot fully capture the intricacies of the continuous action space \citep{lillicrap2015continuous}. Due to this limitation, continuous reinforcement learning approaches are often employed such as proximal policy optimization (PPO) \citep{schulman2017proximal}, to better address the need for precise decision-making in the continuous spectrum of charge and discharge levels.

In continuous reinforcement learning, however, challenges also arise when determining charge and discharge levels due to the dynamic nature of the state of charge (SoC) over time. The range of feasible charge and discharge actions varies based on the evolving SoC. For example, setting actions like charging to a negative value (e.g., complete charging to $-1$) and discharging to a positive value (e.g., complete discharging to $1$) results in a feasible action range of $0$ to $1$ when the battery is fully charged, and $-1$ to $0$ when the battery is fully discharged. Nevertheless, current approaches often struggle to effectively address such time-varying action ranges. Currently, when actions fall outside the designated range, a common solution involves charging up to the maximum SoC or discharging up to the minimum SoC \citep{jeong2021deepcomp,kang2024reinforcement}. However, this approach introduces a potential challenge wherein the SoC may become stuck in a fully charged or fully discharged state during the learning, limiting its ability to explore within the full spectrum of SoC states. Additionally, there are approaches that assign a cost or negative reward proportional to the extent by which actions deviate outside the designated range \citep{lee2020federated,zhang2023safe}. However, there is a potential for overly conservative learning, as the emphasis leans heavily towards actions that remain within the designated range. Addressing these issues is crucial for adapting to the time-varying action ranges associated with the changing SoC over time.

In this paper, we propose a continuous reinforcement learning approach to address these challenges in energy storage device control. The key innovation lies in augmenting the conventional objective functions of the actor and critic with an additional supervising objective function designed to ensure that the output actions at each time step fall within the feasible action space. In contrast to traditional supervised learning, where the training encourages the output to approach specific values, the introduced supervising objective function focuses on constraining the output within a particular range. This inclusion is pivotal as bringing the output within the feasible action space enables the activation of charging and discharging operations in energy storage devices. This proactive approach helps prevent the energy storage from getting stuck in suboptimal states like complete charge or discharge, facilitating the exploration of more optimal actions. The integration of the supervising objective function enhances the adaptability and efficiency of continuous reinforcement learning in optimizing energy storage operations over changing states. We conducted experiments related to energy arbitrage and found that the addition of the supervising objective function effectively addressed the challenge of the system becoming stuck in suboptimal conditions. This objective function prevents the energy storage device from being trapped in states like complete charge or discharge and promotes continual exploration for optimal energy arbitrage.

\section{Methods}
\label{methods}

In this section, we present a continuous reinforcement learning model combined with a supervising objective function. Modern continuous reinforcement learning comprises actor and critic components, each with distinct objective functions depending on the specific reinforcement learning algorithm employed. In this paper, we adopt the proximal policy optimization (PPO) algorithm, known for its compatibility with long short-term memory (LSTM) \citep{schulman2017proximal}, to address energy storage device control problems. When tackling control problems associated with energy storage devices, the majority of reinforcement learning states often involve time-series data such as SoC, energy generation, energy demand, and energy prices. Given this temporal nature, combining PPO with LSTM becomes particularly advantageous. This combination facilitates effective learning and decision-making in scenarios where the state representation is composed of sequential data.

\begin{figure*}[t]
    \centering
    \includegraphics[width=0.8\textwidth]{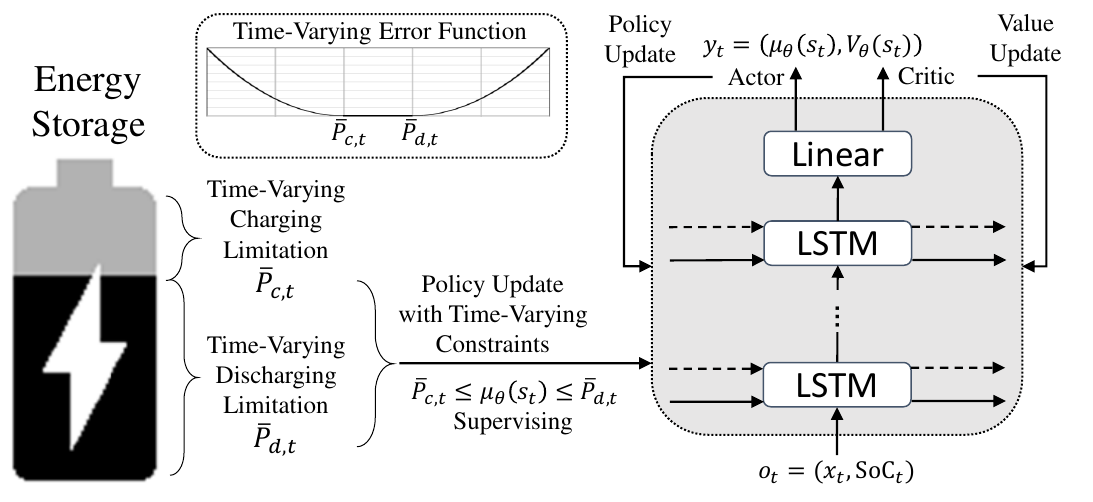}
    \caption{A framework of the proposed method.}\label{fig:frame}
\end{figure*}

The overall framework of the proposed method is shown in Figure~\ref{fig:frame}. The input to the LSTM at each time step is referred to as an observation because, in the context of time-series data, the value corresponding to each time is a partially observable state. At time $t$, the LSTM input, denoted as $o_t$, comprises the SoC at time $t$, represented as $\text{SoC}_t$, and other variables $x_t$ pertinent to the objectives of energy storage device control. These additional variables may include factors such as energy generation, demand, prices, or other relevant parameters, depending on the specific goals of energy storage control. The past observations construct the state at time $t$ as $s_t=\left(o_0, o_1, \cdots, o_t \right)$. In our problem, the action is the charging and discharging amount, and since the action can be known according to changes in the SoC within the state, past actions are not separately included as the state. With the LSTM parameters $\theta$, the output of the LSTM consists of the actor's output, denoted as $\mu_\theta(s_t)$, and the critic's output, denoted as $V_\theta(s_t)$. Here, $\mu_\theta(s_t)$ represents the mean of the Gaussian policy, while $V_\theta(s_t)$ signifies the estimated value. The standard deviation of the Gaussian policy is predetermined based on the desired level of exploration. During the training phase, action $a_t$ is sampled from the Gaussian policy, and during the actual testing phase, $\mu_\theta(s_t)$ serves as the action $a_t$ \citep{zimmer2019exploiting}. Based on the $s_t$ and $a_t$, reward $r_t$ and the next observation $o_{t+1}$ are given from the environment.

The actor and critic objective functions in the standard PPO formulation are expressed as follows:
\begin{equation} \label{eq:actor}
    L^{PPO}_{\text{actor}}(\theta) = \mathbb{E}_t \left[ \min\left(R_t(\theta) \hat{A}_t, \text{clip}(R_t(\theta), 1 - \epsilon, 1 + \epsilon) \hat{A}_t\right)\right],
\end{equation}
\begin{equation} \label{eq:critic}
    L^{PPO}_{\text{critic}}(\theta) = \mathbb{E}_t \left[\left(r_t + \gamma V_\theta(s_{t+1}) - V_\theta(s_t)\right)^2 \right],
\end{equation}
where $R_t(\theta)$ is the ratio of the new and old policies at time $t$, $\hat{A}_t$ is the estimated advantage function at time $t$, $\epsilon$ is a hyperparameter determining the clipping range, and $\gamma$ is a discount factor. We are adding a supervising objective function here. Let the charging limitation at time $t$ as $\bar{P}_{c,t}$ and the discharging limitation as $\bar{P}_{d,t}$. These limitations are determined based on the $\text{SoC}_t$. As the SoC varies over time, both the charging and discharging limitations also time-varying. We have defined charging actions as negative and discharging actions as positive, resulting in $\bar{P}_{c,t} \leq 0$ and $\bar{P}_{d,t} \geq 0$. The proposed supervising objective function is as follows:
\begin{equation} \label{eq:supervising}
    L^{PPO}_{\text{supervising}}(\theta) = \min\left(\mu_\theta(s_t)-\bar{P}_{c,t}, 0 \right)^2 + \min\left(\bar{P}_{d,t}-\mu_\theta(s_t), 0 \right)^2.
\end{equation}
Since $\mu_\theta(s_t)$ serves as the action in the testing phase, we have set the range of $\mu_\theta(s_t)$ to be between $\bar{P}_{c,t}$ and $\bar{P}_{d,t}$. This error function is similar to the mean squared error in supervised learning, with the key distinction that the error is zero within the range of $\bar{P}_{c,t}$ and $\bar{P}_{d,t}$. We finally obtain our main objective, which is minimized at each iteration:
\begin{equation} \label{eq:PPO}
    L^{PPO}(\theta) = L^{PPO}_{\text{actor}}(\theta) + C_1 L^{PPO}_{\text{critic}}(\theta) + C_2 L^{PPO}_{\text{supervising}}(\theta),
\end{equation}
where $C_1$ and $C_2$ are coefficients of the critic objective function and supervising objective function learning, respectively.

\section{Results}
\label{results}

In this section, we evaluate the performance of the proposed method by comparing it with two benchmark cases. Case 1 employed a conventional continuous reinforcement learning approach, excluding the equation (\ref{eq:supervising}), meaning that the output actions were not restricted to be within the feasible action space. Case 2 incorporated the equation (\ref{eq:supervising}) into the reward function. This approach, proposed in \citep{lee2020federated}, adds negative rewards if the output actions fall outside the feasible action space, rather than explicitly learning the range of output actions. The proposed model is designated as Case 3. We demonstrated the effectiveness of the proposed approach through energy arbitrage experiments based on actual energy price data in 2017 U.K. wholesale market \citep{uk2017wholesale}. Accordingly, the additional variable $x_t$ becomes the energy price at time $t$, and the reward $r_t$ is the total profit at time $t$. We take the first 2000 data points which are sampled every 30 minutes and split the dataset into training set (1000 data points), validation set (500 data points), and test set (500 data points) in chronological order, where the validation set was used for early stopping. We normalize the price data between 0 and 1 by the maximum price \$190.81/MWh. We simulate the proposed method using 100MWh battery with the degradation cost of \$10/MWh. At time slot $t=0$, we set $\text{SoC}_t=0.5$, i.e., the half stored energy. We set the minimum and maximum values of the SoC to 0.1 and 0.9, respectively, in order to prevent battery degradation, and the battery charging and discharging model is based on the battery equivalent circuit used in \citep{cao2020deep,jeong2023deep}. We used a 2-layer LSTM architecture with 16 neurons and trained the model using the Adam optimizer. All PPO-related parameters were adopted from values commonly used in \citep{schulman2017proximal}.

\begin{figure}[t]
   \begin{subfigure}{0.3\linewidth}
   \centering
    \includegraphics[width=0.99\linewidth]{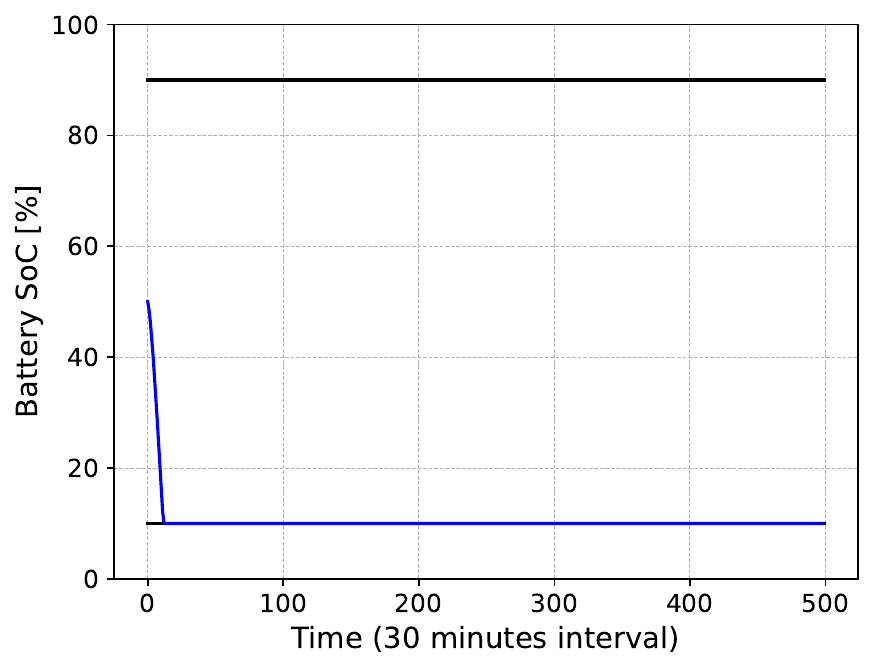}
    \caption{Case 1.}
  \end{subfigure}
   \begin{subfigure}{0.3\linewidth}
   \centering
    \includegraphics[width=0.99\linewidth]{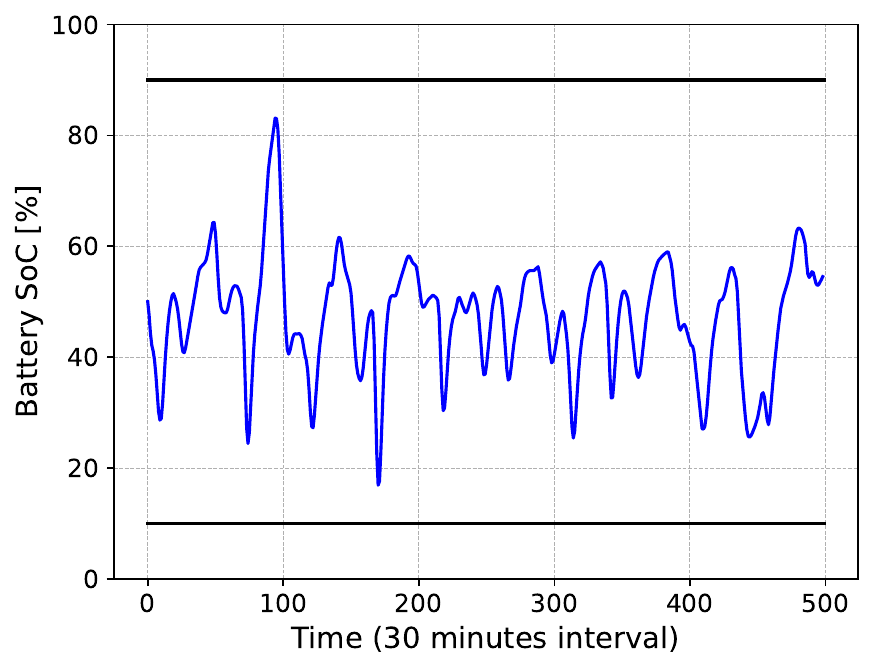}
    \caption{Case 2.}
  \end{subfigure}
   \begin{subfigure}{0.3\linewidth}
   \centering
    \includegraphics[width=0.99\linewidth]{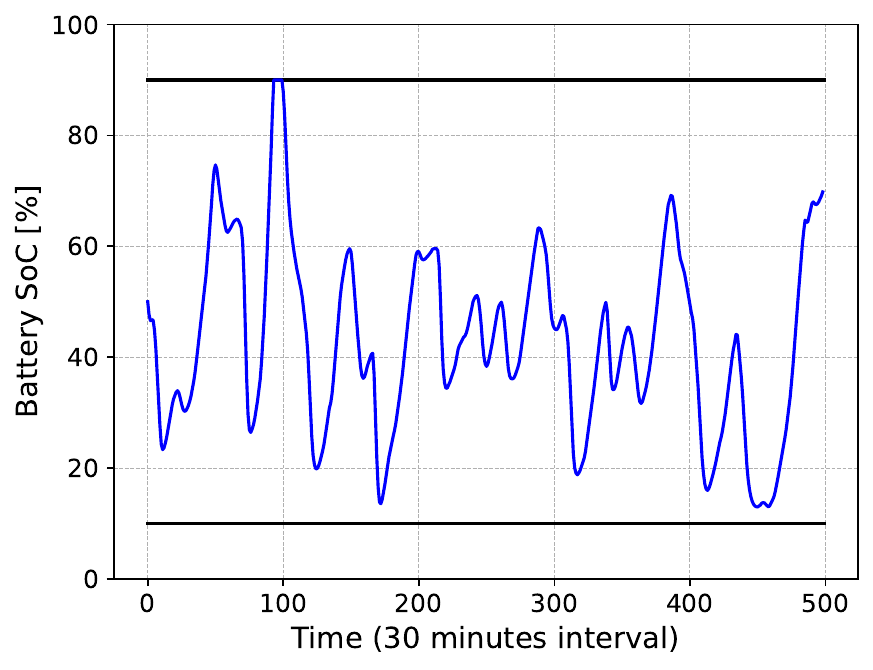}
    \caption{Case 3 (proposed).}
  \end{subfigure}
  \caption{Charging/discharging patterns for 3 cases.}\label{fig_soc}
\end{figure}


\begin{table}[!t]
\caption{Experiment results (30-minutes averaged).} \label{tab:results}
\centering
\begin{normalsize}
\resizebox{0.9\textwidth}{!}{
\begin{tabular}{ccccc}
\toprule
\multirow{2}{*}{} & \multicolumn{4}{c}{Metric} \\ \cmidrule{2-5} 
                  & Charging cost (\$)          & Disharging revenue (\$)        & Degradation cost (\$)          & Total profit (\$)        \\ \midrule
Case 1              & -0.000          & 4.858         & -0.080          & \textbf{4.779}         \\
Case 2              & -40.039       & 53.245      & -1.691       & \textbf{11.514}      \\
Case 3 (proposed)             & -37.493        & 54.085       & -1.714        & \textbf{14.879}       \\ \bottomrule
\end{tabular}
}
\end{normalsize}
\end{table}

Figure~\ref{fig_soc} illustrates the charging and discharging patterns for three cases. Case 1 shows a scenario where all the initially stored energy is discharged (sold out) and no further actions are taken. This suggests a failure in learning to manage the costs associated with charging in energy arbitrage, resulting in suboptimal behavior. Case 2 demonstrates reasonable utilization of energy arbitrage, but the proposed Case 3 engages in more active energy arbitrage. Introducing Equation (\ref{eq:supervising}) as a negative reward makes the agent conservative towards reaching states of complete charge or discharge, leading to reduced utilization of the energy storage. Table~\ref{tab:results} presents the 30-minute average of charging cost, discharging revenue, degradation cost, and total profit for the three cases. It is evident that the proposed Case 3 achieves the highest profit.

\section{Conclusion}
\label{conclusion}

In this paper, we introduce a continuous reinforcement learning approach for energy storage control that considers the dynamically changing feasible charge-discharge range. An additional objective function has been incorporated to learn the feasible action range for each time period. This helps prevent the energy storage from getting stuck in states of complete charge or discharge. Furthermore, the results indicate that supervising the output actions into the feasible action range is more effective in enhancing energy storage utilization than imposing negative rewards when the output actions deviate from the feasible action range. In future research, we will explore combining offline reinforcement learning or multi-agent reinforcement learning to investigate methods for learning a more optimized policy stably.


\subsubsection*{Acknowledgments}
This work was supported by the Korea Institute of Energy Technology and Planning (KETEP) and the Ministry of Trade, Industry \& Energy (MOTIE) of Korea (No. 2021202090028C).

\bibliography{ref}
\bibliographystyle{iclr2024_conference}


\end{document}